\documentclass{article}
\usepackage{spconf,amsmath,graphicx}
\usepackage[T1]{fontenc}
\usepackage{etoolbox}
\usepackage{booktabs}
\usepackage[normalem]{ulem}
\usepackage{multirow}
\usepackage{enumitem}
\usepackage{makecell}
\usepackage{tabularx,ragged2e}

\title{Streaming Anchor Loss: \\Augmenting Supervision with Temporal Significance}

\makeatletter
\def\@name{ \emph{Utkarsh (Oggy) Sarawgi, John Berkowitz, Vineet Garg, Arnav Kundu, Minsik Cho,} \\ \emph{Sai Srujana Buddi, Saurabh Adya, Ahmed Tewfik}\thanks{Thanks to Barry-John Theobald, John Bridle, Pranay Dighe, Ahmed Hussen Abdelaziz, and Mohammad Samragh for feedback.}}
\makeatother
\address{Apple}
\begin{document}
%
\maketitle
\begin{abstract}
Streaming neural network models for fast frame-wise responses to various speech and sensory signals are widely adopted on resource-constrained platforms. Hence, increasing the learning capacity of such streaming models (i.e., by adding more parameters) to improve the predictive power may not be viable for real-world tasks. In this work, we propose a new loss, \textit{Streaming Anchor Loss (SAL)}, to better utilize the given learning capacity by encouraging the model to learn more from essential frames. More specifically, our SAL and its focal variations dynamically modulate the frame-wise cross entropy loss based on the importance of the corresponding frames so that a higher loss penalty is assigned for frames within the temporal proximity of semantically critical events. Therefore, our loss ensures that the model training focuses on predicting the relatively rare but task-relevant frames. Experimental results with standard lightweight convolutional and recurrent streaming networks on three different speech based detection tasks demonstrate that SAL enables the model to learn the overall task more effectively with improved accuracy and latency, without any additional data, model parameters, or architectural changes. 
\end{abstract}
\section{Introduction}\label{sec:intro}
Many machine learning systems operate on continuous time-series data streams to detect and predict target variables \cite{gomes2019machine, sarawgi2021uncertainty, zhao2019raise, zhang2022wakeupnet, higuchi20_interspeech, sarawgi20_interspeech}. Improving the performance of streaming models has typically involved increasing the number of parameters, finding better architectures or adding more training data \cite{gomes2019machine, zhang2022wakeupnet, garg21_interspeech}. These approaches, while effective, come with crucial real-world challenges, including increased requirements and complexity of hardware, higher latency, and unavailability of large-scale representative data. Also, adding more data is not always helpful for small models with limited capacity \cite{shrivastava2021optimize}.

Streaming models often use connectionist temporal classification (CTC) or recurrent neural network transducer (RNN-T) losses for alignment-free training \cite{graves2006connectionist, rao2017exploring, zhang2020transformer, garg21_interspeech, graves2012sequence} and frame-wise cross entropy or focal losses for frame-wise training \cite{zhang2022wakeupnet, shrivastava2021optimize, kim2022ada, buddi2023efficient, xu2021lightweight, dinkel2021towards}. Frame-wise cross entropy loss (FCEL) computes the cross entropy loss between the frame prediction and the frame ground truth label, which can be sub-optimal in streaming tasks as the training occurs on a highly imbalanced set of frames with respect to the prediction class as well as the importance of the frames in the task \cite{zhang2022wakeupnet, shrivastava2021optimize, kim2022ada, higuchi2021dynamic, du2022efficient, ishikawa2021alleviating}. Saxena et al. \cite{saxena2019data} and Higuchi et al. \cite{higuchi2021dynamic} use learnable data parameters that modulate the importance of data samples and classes during training. Ryou et al. \cite{ryou2019anchor} introduced an anchor loss, improving over focal loss \cite{lin2017focal}, that dynamically rescales the cross entropy in image classification and human pose estimation based on prediction difficulty for samples. Some image and video segmentation works modify the regular pixel-wise and frame-wise cross entropy to better learn around the boundary pixels and frames respectively, which are usually more challenging \cite{du2022efficient, ishikawa2021alleviating, chen2020contour, borse2021inverseform}. 

\begin{figure}[htb]
\begin{minipage}[b]{0.48\linewidth}
  \centering
  \centerline{\includegraphics[width=4.0cm]{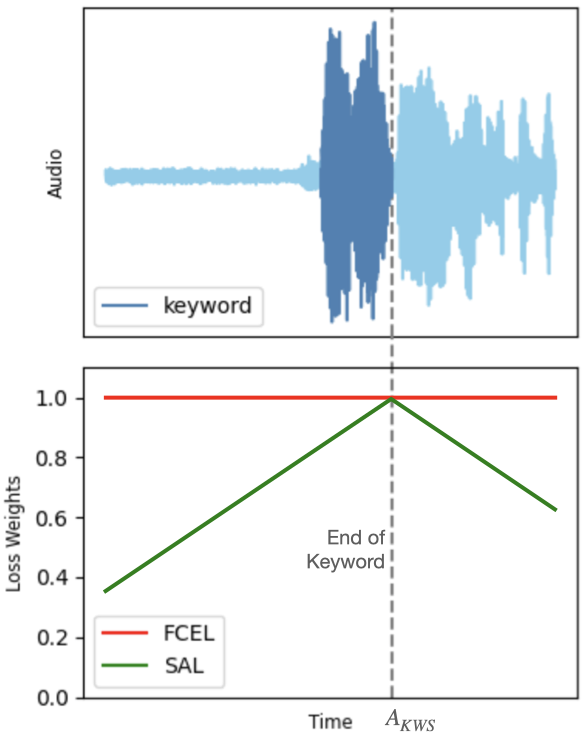}}
  \centerline{(a) Keyword Spotting}\medskip
\end{minipage}
\hfill
\begin{minipage}[b]{0.48\linewidth}
  \centering
  \centerline{\includegraphics[width=4.0cm]{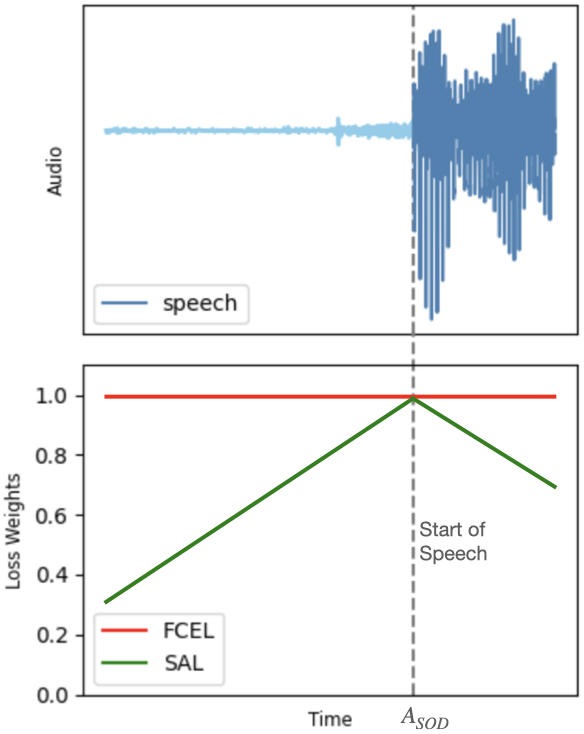}}
  \centerline{(b) Speech Onset Detection}\medskip
\end{minipage}
\caption{Toy examples for (a) Keyword Spotting (KWS) and (b) Speech Onset Detection (SOD) showing the corresponding task anchors, $A_{KWS}$ and $A_{SOD}$, respectively (Section \ref{sec:anchor}), and how the streaming anchor loss (SAL) modulates the frame-wise cross entropy loss (FCEL) with a frame-wise multiplicative factor $w_t$, referred to as loss weights in the plots above, such that $l_{SAL} = w_t * l_{FCEL}$ ($w_t$ in green) and $l_{FCEL} = w_t * l_{FCEL}$ ($w_t$ in red).}
\label{fig:loss_plots}
\end{figure}

We propose a new loss for training streaming models to learn corresponding frame-wise predictions in speech and sensory based tasks without modifying the resource-efficient model architecture or increasing the required training data. Specifically, we introduce a frame-wise loss function, \textit{streaming anchor loss (SAL)}, that is temporally more informed of the time-series task and corresponding labels, enabling the model to learn the prediction for each frame or timestep with varying importance. The loss incorporates task-specific domain knowledge into the loss using \textit{anchors}, that are the frames crucial to the detection task's ground truth labels. Instead of gathering new training data, which is often expensive and time consuming, our method augments the existing training data with anchor points, allowing SAL to use the available data more effectively. SAL dynamically rescales the frame-wise cross entropy based on the temporal proximity to task-specific anchors. Figure \ref{fig:loss_plots} illustrates examples in the context of particular streaming tasks where the loss function penalizes errors on frames closer to the task anchors, i.e., the crucial frames, more heavily
(detailed elaborations in Section \ref{sec:anchor}). Such a loss regularizes the model training to prioritize its limited available resources on predicting the infrequent and significant frames primarily responsible for the task at hand. Experimental results demonstrate that the loss significantly improves the performance across different model architectures, detection tasks, and data modalities. 

To this end, we make the following contributions:
\vspace{-0.25mm}
\begin{itemize}
    \item We propose a new streaming anchor loss and its focal variations to learn frame-wise predictions for streaming tasks. We present a general formulation of the losses and adopt them for real-world tasks (Section \ref{sec:methods}).
    \item We demonstrate generalizability and effectiveness of the proposed losses using common resource efficient model architectures like convolutional and recurrent networks on three specific tasks using audio and multi-modal data - keyword spotting, multi-modal trigger detection, and speech onset detection.
    \item Compared to baseline frame-wise losses, our proposed losses achieve improved accuracy as well as detection latency (Section \ref{sec:results}).
\end{itemize}

\section{Methods and Materials}\label{sec:methods}
\subsection{Notation and Setup}
\label{sec:setup}
In dataset $D$ consisting of $N$ datapoints, we define the $n^{th}$ datapoint $D^{(n)} \equiv \{X_t^{(n)}, y_t^{(n)}\}_{t=1}^{T_n}$, where $T_n$ is the total number of timesteps or frames, and $X_t^{(n)}$ and $y_t^{(n)} \in \{0,1\}$ are the input data and the binary scalar ground truth values respectively at timestep $t$ of the $n^{th}$ datapoint. Let $\hat{y}_t^{(n)} \in [0,1]$ be the model's prediction of positive class at timestep $t$ for the $n^{th}$ datapoint.  We henceforth focus on a single datapoint and drop the dependence on $n$. We then have the total loss between ground truth $y$ and prediction $\hat{y}$ as the frame-wise loss $l(y_t, \hat{y}_t)$ averaged over all frames and all datapoints. All tasks discussed in this paper use binary labels, but the methods developed can be generalized to multi-class tasks.

\subsection{Streaming Anchor Loss}
\label{sec:sal}
Streaming anchor loss (SAL) uses a frame-wise weight, $w_t$, to modulate the frame-wise cross-entropy loss (FCEL) such that this weight contains information about the importance of the corresponding frames in context of the task. Scaling the frame-wise loss by this weight makes the model learn the task better by penalizing the errors more on crucial frames. The importance of the frames is defined with respect to what we call \textit{task anchors}, $A_{task} \in \{1,...,T\}$. Frames temporally closer to the task anchor frames are considered more important and thus assigned higher weight compared to the ones further away from the anchors, as formulated in Equation \eqref{eq:sal_loss}, with the task anchor frames given the largest weight. This allows the model to learn and predict the crucial frames better, thus improving the overall performance in the streaming task.

Let $l_{FCEL}(y_t, \hat{y}_t)$ and $l_{SAL}(y_t, \hat{y}_t)$ respectively denote the frame-wise cross entropy loss and streaming anchor loss. \textit{Frame-wise Cross Entropy Loss (FCEL)}, $l_{FCEL}(y_t, \hat{y}_t)$, is defined as follows:
\begin{equation}\label{eq:cross_entropy_loss}
l_{FCEL}(y_t, \hat{y}_t) = -(y_t) *  log(\hat{y}_t) - (1-y_t) * log(1 - \hat{y}_t)
\end{equation}
We define the \textit{Streaming Anchor Loss (SAL)}, $l_{SAL}(y_t, \hat{y}_t)$, as follows:
\begin{equation} \label{eq:sal_loss}
\begin{split}
l_{SAL}(y_t, \hat{y}_t) & = w_t * l_{FCEL}(y_t, \hat{y}_t)\\
\text{where,}\hspace{10mm} w_t & = \frac{T - |A_{task} - t|}{T}
\end{split}
\end{equation}
At the anchor frame, i.e. for $t = A_{task}$, $w_t=1=w_{A_{task}}$, and $w_t$ decreases linearly as $t$ moves further from $A_{task}$. For a datapoint with no event, and thus no anchor frame, we weight all frames uniformly. Note that there is no more than one $A_{task}$ per datapoint, as also elaborated for all tasks in Section \ref{sec:anchor}, but the methods developed can be extended to data with multiple events. 

\subsubsection{Focal variations of Streaming Anchor Loss}
Recent works \cite{zhang2022wakeupnet, kim2022ada} have used a temporal adaptation of focal loss \cite{lin2017focal}, what we refer to as frame-wise focal loss (FFL), to learn frame-wise predictions since there is usually a class imbalance as most frames belong to background (trigger absent, speech absent, etc) and significantly fewer frames belong to foreground (trigger present, speech present, etc). As such, we compare our proposed SAL with FFL and also propose two focal variations of streaming anchor loss: streaming anchor+focal loss (SA+FL) and streaming anchor focal loss (SAFL). We summarize the intuitive difference between these losses and mathematically formulate them below.

\textit{Frame-wise Focal Loss (FFL)}: The loss is modified for challenging frames dependent on the classifier confidence \cite{lin2017focal}. This is mathematically formulated as $l_{FFL}(y_t, \hat{y}_t)$ in Equation \eqref{eq:ffl} below. We use $\gamma = 2$ and $\alpha = 0.25$ as per our best FFL results and Lin et al. \cite{lin2017focal}.
\begin{equation}\label{eq:ffl} \begin{split}
l_{FFL}(y_t, \hat{y}_t) = - (y_t) * \alpha * F_{\gamma}(\hat{y}_t)\\
-  (1-y_t) * (1-\alpha) * F_{\gamma}(1 - \hat{y}_t)\\
\text{where,}\hspace{5mm}F_{\gamma}(s) = (1-s)^{\gamma}*log(s) 
\end{split} \end{equation}
\textit{Streaming Anchor+Focal Loss (SA+FL)}: The model learns by relatively up-weighting the loss for frames temporally closer to the task anchors, or for challenging frames dependent on the classifier confidence. We define this loss as $l_{SA+FL}(y_t, \hat{y}_t)$ in Equation \eqref{eq:sa+fl} below.
\begin{equation}\label{eq:sa+fl}
    l_{SA+FL}(y_t, \hat{y}_t) = l_{SAL}(y_t, \hat{y}_t) + l_{FFL}(y_t, \hat{y}_t)
\end{equation}
\textit{Streaming Anchor Focal Loss (SAFL)}: The model learns by relatively up-weighting the loss for the challenging frames which are themselves temporally closer to the task anchors. As compared to SA+FL, SAFL assigns higher penalty specifically for frames that are both temporally closer to the task anchors as well as challenging based on classifier confidence. Similar to Equation \eqref{eq:sal_loss}, we define this loss as $l_{SAFL}(y_t, \hat{y}_t)$ in Equation \eqref{eq:safl} below.
\begin{equation}\label{eq:safl}
    l_{SAFL}(y_t, \hat{y}_t) = w_t * l_{FFL}(y_t, \hat{y}_t)
\end{equation}

\subsubsection{Anchors for different streaming tasks}\label{sec:anchor}
In this section, we define and adopt task anchors for training streaming models on different tasks on which we evaluate. Figure \ref{fig:loss_plots} illustrates the same with toy examples on keyword spotting (KWS) and speech onset detection (SOD) tasks.

\vspace{2mm}\textit{Keyword Spotting (KWS)}: KWS model is streaming by nature so as to detect and localize the presence of a keyword in the input audio. The model should trigger when the keyword has ended. As such, we define the anchor for the KWS task, $A_{KWS}$, as the frame corresponding to the end of keyword, using the ground truth labels.

\vspace{2mm}\textit{Multi-modal Trigger Detection (MTD)}: Recent works have introduced and developed more natural and convenient ways of interacting with voice assistants and smartwatches using gesture, i.e. accelerometer data, and speech i.e. audio \cite{zhao2019raise, buddi2023efficient}. Similar to a KWS model, MTD model is streaming in nature so as to detect and localize the presence of a trigger in the multi-modal input data. MTD allows a user to simply raise their watch to their mouth and speak, without the need for any explicit keyword. The model should trigger when speech has started while the raising gesture is either in progress or has ended. As such, we define the anchor for the MTD task, $A_{MTD}$, as the frame corresponding to the start of speech, using the ground truth labels for both modalities.

\vspace{2mm}\textit{Speech Onset Detection (SOD)}: SOD model is streaming by nature so as to detect and localize the onset, i.e. start of speech. The model should trigger when the speech has started. As such, we define the anchor for the SOD task, $A_{SOD}$, as the frame corresponding to the start of speech, using the ground truth labels.

\vspace{-1mm}
\section{Experiments and Results}\label{sec:results}
\subsection{Keyword Spotting}\label{sec:kws-results}
\vspace{-1mm}
Similar to previous works \cite{higuchi20_interspeech, shrivastava2021optimize}, our KWS dataset contains a total of $500$k near and far-field utterances (captured at $3$ft and $6$ft distance) with the keyword ``Siri'' followed by a user query and additional $500$k utterances without a keyword. We augment all of the data using room-impulse responses (RIRs), echo residuals, and ambient noise samples from various acoustic environments, as well as apply gain augmentation of $10$dB to $-40$dB. We use predefined splits of train, validation, and test sets with ratios of $70\%$, $15\%$, and $15\%$, respectively.

For a resource efficient setup, we extract $16$ dimensional MFCC features from audio that are fed into a $12$k parameter convolutional model (CNN). The lightweight model consists of $14$ $1$-D convolutional blocks ($1$ depthwise $+$ $3\times$($2$ depthwise $+$ $1$ pointwise block) $+$ pointwise output layer) with an effective receptive field of $153$. We train the model with aligned frame-wise binary labels with an Adam optimizer and a learning rate scheduler using cosine annealing with an initial learning rate of $0.005$. All experiments use the same setup except different training loss functions, for a fair comparison. 

We evaluate the models on a threshold-independent metric, AUC ROC, and keyword spotting latency, defined as the absolute time difference between the ground truth keyword detection, i.e. the end of the labeled keyword frames, and the predicted keyword detection. Table \ref{tab:kws-results} shows the results across different loss functions. We observe that SAL and SAFL improve AUC ROC and yield up to $48.9\%$ relative improvement in mean latency.

\vspace{-2mm}
\begin{table}[h]
  \caption{Keyword spotting results comparing the different losses with the streaming CNN model using audio.}
  \label{tab:kws-results}
  \centering
  \vspace{2mm}
  \begin{tabular}{lcc}
    \toprule
      Loss & AUC ROC ($\%$) & Mean Latency (secs)\\
       \toprule
       FCEL & 97.61 & 1.35\\
       FFL & 98.53 & 0.70\\
       SAL & \textbf{99.53} & \textbf{0.69}\\
       SA+FL & 97.75 & 0.76\\
       SAFL & 98.97 & 0.70\\
    \bottomrule
  \end{tabular}
\end{table}

\vspace{-3mm}
\subsection{Multi-modal Trigger Detection}\label{sec:mmi-results}
Similar to previous works \cite{zhao2019raise, buddi2023efficient}, the MTD dataset contains a total of 8500 hours of multi-modal data from 2000 subjects. Audio recordings from the single-channel $16$ kHz microphone and $3-$axis accelerometer data were collected using a smartwatch through internal user studies with informed consent approvals. We use predefined splits of train, validation, and test sets with ratios of $70\%$, $15\%$, and $15\%$, respectively.

We follow the previous works \cite{zhao2019raise, buddi2023efficient} for a fair comparison and to use a resource efficient system. The system consists of three different streaming models $-$ audio model, gesture model, and fusion model. We use the same audio and gesture features and models as \cite{zhao2019raise, buddi2023efficient}. The frame-wise outputs of the audio and gesture models feed into a streaming fusion model. We use the best performing fusion model and same training hyperparameters from \cite{buddi2023efficient}. The lightweight fusion model is a Gated Recurrent Unit (GRU) model with $13$k parameters \cite{buddi2023efficient}. All experiments use the same setup except the different training loss functions, for a fair comparison.

To estimate the quality of such detection systems, we evaluate the model's false positive rate (FPR) and false negative rate (FNR).  A session is labeled positive if and only if it contains at least one frame labeled as containing a trigger, and is given a positive prediction if the model score exceeds the tuned threshold on at least one frame in the session. We define trigger detection latency as the absolute time difference between the ground truth trigger detection, i.e. the start of the labeled trigger frames, and the predicted trigger detection. 
Similarly, Brier scores \cite{Brier1950VERIFICATIONOF} are computed using the maximum score across the session. For a fair comparison between the models, we report FNR at a fixed $2\%$ FPR, and use the corresponding thresholds tuned to that operating point. The presented trends hold true across a variety of FPR operating points. Table \ref{tab:mmi-results} shows the results across different loss functions. We observe that SAL, SA+FL, and SAFL yield relative improvements up to $52.1\%$ in FNR/FPR, $18.4\%$ in mean latency, and $37.2\%$ in Brier score. 
\vspace{-2mm}
\begin{table}[h]
  \caption{Multi-modal trigger detection results comparing the different losses with the streaming GRU model using fused modalities - gesture and audio. $*$ represents the baseline \cite{buddi2023efficient} retrained with the same data as other competing models.}
  \label{tab:mmi-results}
  \centering
  \vspace{2mm}
  \begin{tabular}{lccc}
    \toprule
      \multirow{2}{*}{Loss} & \% FNR & Mean Latency & Brier\\
      & @ 2\% FPR & (millisecs) & Score (\%)\\
       \toprule
       FCEL $*$ & 11.06 & 369 & 5.83\\
       FFL & 6.66 & 312 & 3.87\\
       SAL & 5.50 & 310 & 3.97\\
       SA+FL & 5.40 & 305 & \textbf{3.66}\\
       SAFL & \textbf{5.30} & \textbf{301} & 3.77\\
    \bottomrule
    \vspace{-2mm}
  \end{tabular}
\end{table}

\vspace{-5mm}
\subsection{Speech Onset Detection}\label{sec:sod-results}
We use the same audio recordings and the corresponding speech onset labels as in Section \ref{sec:mmi-results} with predefined splits of train, validation, and test sets with ratios of $70\%$, $15\%$, and $15\%$, respectively. We extract $40$ dimensional mel filter banks from audio that are fed into a $90$k parameters model. The model consists of an LSTM layer with $128$ hidden size, followed by $2$ dense layers with $0.2$ dropout projecting the $128$-dim LSTM outputs to $32$ and $2$ respectively. We train the model with an Adam optimizer and a learning rate of $0.001$. All experiments use the same setup except the different training loss functions, for a fair comparison. 

We use the same setup as Section \ref{sec:mmi-results} for measuring latency, herein defined as the absolute time difference between the predicted speech onset and the ground truth speech onset for this task.  Table \ref{tab:sod-results} collates some descriptive statistics of the resultant latency distributions across different loss functions.  We observe that SAL helps improve over the baselines, and SA+FL and SAFL in particular demonstrate marked improvements across different quantiles of the latency distribution.
\vspace{-1mm}
\begin{table}[h]
  \caption{Speech onset detection results comparing the different losses with the streaming LSTM model using audio.}
  \label{tab:sod-results}
  \centering
  \vspace{2mm}
  \setlength{\tabcolsep}{12pt}
  \begin{tabular}{lcccc}
    \toprule
      \multirow{2}{*}{Loss} & \multicolumn{4}{c}{Latency (secs)}\\
      \cmidrule{2-5}
      & Mean & p25 & p50 & p75\\
       \toprule
       FCEL & 2.64 & 2.23 & 2.61 & 2.99\\
       FFL & 2.61 & 2.21 & 2.60 & 2.99\\
       SAL & 2.52 & 2.10 & 2.56 & 2.96\\
       SA+FL & \textbf{1.14} & \textbf{0.46} & \textbf{0.86} & \textbf{1.26}\\
       SAFL & 1.37 & 0.65 & 1.05 & 1.53\\
    \bottomrule
  \end{tabular}
\end{table}
\vspace{-5mm}
\section{Discussion and Future Work}\label{sec:discuss}
We proposed a new family of loss functions for streaming classification based upon increasing the importance of frames near the primary events that define the task.  We have demonstrated that this loss function improves accuracy and predictive latency across multiple tasks and standard model architectures. The reduced latency also makes it valuable for practical implementation in real-time and time-sensitive tasks. Resource-constrained models which are central to such tasks particularly show more pronounced effects, as we believe the loss function regularizes the model during training to use most of its limited capacity on classifying the relatively rare and task-relevant anchor frames. It is also intuitive that improved prediction of such anchor frames would not only improve accuracy of overall task but also reduce latency since these frames signify the primary predictive event of the task.

While the examples explored in this paper focus on binary detection tasks, future work will extend this framework to more general multi-class classification streaming tasks with more complex notions of anchor frames. It is also difficult to precisely locate the start or end of relevant segments either due to inter-annotator disagreement or the use of an automated data collection protocol \cite{xu2021lightweight}, so adapting these loss functions to a weakly-supervised paradigm will be helpful.

\bibliographystyle{main}
\bibliography{main}

\end{document}